\documentclass[conference]{IEEEtran}
\IEEEoverridecommandlockouts

\usepackage{cite}
\usepackage{amsmath,amssymb,amsfonts}
\usepackage{graphicx}
\usepackage{textcomp}
\usepackage{xcolor}
\usepackage{colortbl}
\usepackage{multirow}
\usepackage{booktabs}
\usepackage{array}
\usepackage{siunitx}
\usepackage{balance}
\usepackage{orcidlink}
\definecolor{rowbest}{RGB}{198,239,206}   % soft green  -- best result
\definecolor{rowgood}{RGB}{255,242,204}   % soft amber  -- notable / viable
\definecolor{rowbad} {RGB}{255,212,211}   % soft red    -- failure mode
\definecolor{celltop}{RGB}{169,221,178}   % deeper green -- single best cell

\def\BibTeX{{\rm B\kern-.05em{\sc i\kern-.025em b}\kern-.08em
    T\kern-.1667em\lower.7ex\hbox{E}\kern-.125emX}}
\graphicspath{{figs/}}

\title{Unsupervised LiDAR-Based Multi-UAV Detection and Tracking Under Extreme Sparsity}

\author{
\IEEEauthorblockN{1\textsuperscript{st} Nivand Khosravi\,\orcidlink{0009-0004-3749-7249}}
\IEEEauthorblockA{\textit{Instituto Superior T\'{e}cnico} \\
\textit{University of Lisbon} \\
Lisbon, Portugal \\
nivand.khosravi@tecnico.ulisboa.pt}
\and
\IEEEauthorblockN{2\textsuperscript{nd} Rodrigo Ventura\,\orcidlink{0000-0002-5655-9562}}
\IEEEauthorblockA{\textit{Instituto Superior T\'{e}cnico} \\
\textit{University of Lisbon} \\
Lisbon, Portugal \\
rodrigo.ventura@tecnico.ulisboa.pt}
\and
\IEEEauthorblockN{3\textsuperscript{rd} Meysam Basiri\,\orcidlink{0000-0002-8456-6284}}
\IEEEauthorblockA{\textit{Instituto Superior T\'{e}cnico} \\
\textit{University of Lisbon} \\
Lisbon, Portugal \\
meysam.basiri@tecnico.ulisboa.pt}
}

\begin{document}

\maketitle

\begin{abstract}
Non-repetitive solid-state LiDAR scanning leads to an extremely sparse
measurement regime for detecting airborne UAVs: a small quadrotor at
10--25\,m typically produces only 1--2 returns per scan, which is far below the
point densities assumed by most existing detection approaches and inadequate
for robust multi-target data association. We introduce an unsupervised,
LiDAR-only pipeline that addresses both detection and tracking without the
need for labeled training data. The detector integrates range-adaptive DBSCAN
clustering with a three-stage temporal consistency check and is benchmarked
on real-world air-to-air flight data under eight different parameter
configurations. The best setup attains 0.891 precision, 0.804 recall, and
0.63\,m RMSE, and a systematic minPts sweep verifies that most scans contain at
most 1--2 target points, directly quantifying the sparsity regime. For
multi-target tracking, we compare deterministic Hungarian assignment with joint probabilistic data association (JPDA), each coupled with
Interacting Multiple Model filtering, in four simulated scenarios with
increasing levels of ambiguity. JPDA cuts identity switches by 64\% with
negligible impact on MOTA, demonstrating that probabilistic association is
advantageous when UAV trajectories approach one another closely. A
two-environment evaluation strategy, combining real-world detection with
RTK-GPS ground truth and simulation-based tracking with identity-annotated
ground truth, overcomes the limitations of GNSS-only evaluation at
inter-UAV distances below 2\,m.
\end{abstract}

%=============================================================================
\section{Introduction}
\label{sec:intro}
%=============================================================================

Growing deployment of small UAVs in civilian and security-sensitive airspace
demands onboard perception systems that detect aerial targets and maintain
consistent track identities. In multi-UAV operations, missed detections and
identity switches corrupt relative state estimates, undermining collision
avoidance, formation control, and cooperative
autonomy~\cite{ritchie2015micro,nguyen2018drone}. Reliable airborne perception
is therefore a foundational requirement for autonomous multi-UAV missions,
including leader-follower navigation, coordinated inspection, and swarm
operations in GPS-degraded environments.

Among sensing modalities, RF~\cite{ezuma2019detection} is susceptible to
interference; acoustic methods~\cite{basiri2016sound} degrade outdoors;
vision-based approaches~\cite{rozantsev2017detecting} require adequate
lighting and appearance cues; and radar exhibits elevated false alarm rates
for small, low-RCS targets~\cite{bjorklund2018mutual}. LiDAR provides
illumination-independent 3D geometry~\cite{hammer2018lidar} independent of
target appearance, making it attractive for onboard aerial perception.

Solid-state LiDAR with non-repetitive scanning creates a perception regime
fundamentally different from ground robotics. At 10--25\,m, a small quadrotor
produces only 1--2 returns per scan due to the rosette pattern that
concentrates points non-uniformly each cycle. This extreme sparsity makes
fixed-parameter DBSCAN brittle: targets fall below minimum-point thresholds
in consecutive scans. In multi-UAV scenarios, overlapping returns from two
targets within 1--2\,m cannot be reliably disambiguated by deterministic
assignment.

Addressing this requires a detector and tracker co-designed for the sparsity
regime. The detector must tolerate range-dependent density variation while
rejecting background clutter, and the tracker must preserve target identities
through trajectory crossings and prolonged detection outages. Per-frame Hungarian
matching~\cite{kuhn1955hungarian} is efficient but commits irrevocably to a
single hypothesis per step, accumulating identity switches under proximity
ambiguity. Joint Probabilistic Data
Association~\cite{bar1988tracking,fortmann1983sonar,rezatofighi2015joint}
distributes likelihood across all feasible hypotheses, offering inherent
robustness to measurement ambiguity. IMM
filtering~\cite{blom1988imm,li2003survey} handles maneuvering 3D motion via
adaptive submodels. Maintaining continuity through LiDAR dropout via adaptive
EKF with motion-model
switching~\cite{khosravi2026lightweight3dlidarbaseduav} is a complementary
strategy targeting the tracking stage. In contrast, the present work embeds
temporal consistency within the detector and provides a controlled comparison
of deterministic and probabilistic association for sparse LiDAR multi-UAV
tracking.
The main contributions of this work are as follows:
\begin{enumerate}
\item \textbf{An unsupervised LiDAR-based detector for extreme sparsity.}
We develop a range-adaptive DBSCAN detector with three-stage temporal
consistency validation that operates without labeled training data. Evaluated
across eight real-world configurations, the best setting achieves 0.891
precision, 0.804 recall, and 0.63\,m RMSE. A controlled minPts sweep further
shows that most target observations contain only 1--2 points per scan,
directly characterizing the underlying sparsity regime.

\item \textbf{A comparative study of data association for sparse multi-UAV tracking.}
We compare deterministic Hungarian assignment and probabilistic JPDA within an
IMM-based tracking framework across four simulation scenarios of increasing
association ambiguity. The results show that JPDA reduces identity switches by
64\%, from 4.4 to 1.6 on average, at a MOTA cost of only 0.003.

\item \textbf{A two-environment evaluation protocol for detection and tracking.}
We combine real-world air-to-air detection experiments, using RTK-GPS ground
truth, with simulation-based tracking experiments that provide unambiguous
identity labels at sub-2\,m inter-UAV separations. This protocol enables
rigorous evaluation in conditions where GNSS-only ground truth is insufficient.
\end{enumerate}
%=============================================================================
\section{Related Work}
\label{sec:related}
%=============================================================================

\subsection{LiDAR-Based Aerial Target Detection}

Hammer et al.~\cite{hammer2018lidar} established detection feasibility for
sub-0.5\,m quadrotors within 30\,m using spinning LiDAR, noting rapid
point-count decline with range. Razlaw et al.~\cite{razlaw2019detection}
applied temporal integration for cluttered urban environments. Dogru and
Marques~\cite{dogru2022drone} demonstrated detection with only 3--5 points
per target using spatial-temporal features, and Liang
et al.~\cite{liang2024unsupervised} achieved unsupervised trajectory
estimation without labeled training data.

These works assume spinning LiDAR with repeatable scan patterns and several
points per scan. Non-repetitive solid-state sensors such as the Livox Mid-360
use rosette scanning that repositions coverage each cycle, pushing targets to
1--2 points at practical ranges, where fixed clustering fails and cross-frame
temporal integration becomes essential.

Learning-based 3D detection architectures including
PointNet~\cite{qi2017pointnet}, PointNet++~\cite{qi2017pointnetpp},
VoxelNet~\cite{zhou2018voxelnet}, PointPillars~\cite{lang2019pointpillars},
and PointRCNN~\cite{shi2019pointrcnn} perform strongly on automotive datasets
where targets provide dozens to hundreds of points. These architectures
implicitly assume consistent training and test point-density distributions;
under extreme sparsity, input distributions shift severely and learned
geometric features lose discriminative power. Unsupervised clustering
avoids this sensitivity, and range-adaptive
DBSCAN~\cite{ester1996dbscan} naturally compensates for density decay with
range without labeled data, a key advantage for resource-constrained
deployment.

\subsection{Data Association for Multi-Target Tracking}

AB3DMOT~\cite{weng2020ab3dmot}, SORT~\cite{bewley2016simple}, and Deep
SORT~\cite{wojke2017simple} show Hungarian matching is effective in automotive
3D tracking where detections are dense and targets well-separated. Chiu
et al.~\cite{chiu2021probabilistic} demonstrated that explicit association
uncertainty modeling improves identity consistency in ambiguous scenarios.

Airborne LiDAR tracking differs fundamentally: full-3D maneuvering at higher
agility, low detection rates from extreme sparsity, and trajectory crossings
as the dominant error source mean standard automotive trackers are unsuitable.
We compare Hungarian and JPDA at the tracker level using
IMM~\cite{blom1988imm,li2003survey} under controlled scenarios and provide an
explicit identity-switch analysis, a comparison not previously reported for
this setting.
%=============================================================================
\section{Methodology}
\label{sec:methodology}
%=============================================================================

The system comprises an unsupervised LiDAR detector producing sparse 3D
measurements and a multi-object tracker estimating continuous target
trajectories using IMM filtering with alternative data-association strategies.
Detection operates in the local UAV frame~$\mathcal{F}_L$ for geometric
filtering; tracking operates in the global inertial frame~$\mathcal{F}_G$
to keep constant-velocity motion models valid under ego-motion. The modular
architecture allows detection configurations and association strategies to be
varied independently.

\subsection{Coordinate Frames and Notation}

The local frame~$\mathcal{F}_L$ is body-fixed on the observer UAV; the global
frame~$\mathcal{F}_G$ is a world-fixed inertial reference established at
initialization. Raw points are processed in~$\mathcal{F}_L$ because geometric
constraints (height bounds, radial range limits) are natural in the sensor
frame. All tracking operations are performed in~$\mathcal{F}_G$, ensuring
constant-velocity models remain valid under arbitrary ego-rotations and
translations.

At discrete time~$t_k$, the LiDAR scan is an unordered point set
$\mathcal{P}_k = \{\mathbf{p}_k^{(n)} \in \mathbb{R}^3\}_{n=1}^{N_k}$
in~$\mathcal{F}_L$, where $N_k$ is the valid return count. The detector
outputs $m_k$ candidate measurements
$\mathcal{Z}_k = \{\mathbf{z}_{k,j}\}_{j=1}^{m_k}$ as cluster centroids
in~$\mathcal{F}_L$, transformed to~$\mathcal{F}_G$ via the known ego-pose
before being passed to the tracker.

\subsection{Unsupervised LiDAR-Based UAV Detection}

The detection pipeline extracts validated target measurements through ROI
filtering and voxel downsampling, range-adaptive DBSCAN clustering, and
three-stage temporal consistency validation, each stage operating under the 1--2
point-per-scan constraint without labeled training data.

\subsubsection{Preprocessing and Range-Adaptive Clustering}

Raw clouds $\mathcal{P}_k$ undergo region-of-interest (ROI) filtering that
removes ground returns below height $h_{\min}$ relative to the sensor, discards
far-field points beyond $r_{\max}$, and excludes self-returns within a
cylindrical exclusion zone of radius $r_{\mathrm{excl}}$ around the sensor
origin to prevent observer-platform interference. Voxel downsampling via the
Point Cloud Library~\cite{rusu20113d,holz2015registration} is then applied:
\begin{equation}
\mathcal{P}_k^{V} = \mathrm{VoxelGrid}(\mathcal{P}_k^{\mathrm{ROI}};\, v),
\end{equation}
where voxel size $v$ controls the spatial resolution--computation trade-off.
Each occupied voxel is represented by the centroid of its contained points,
retaining geometric fidelity at reduced cardinality~\cite{rusu20113d}.

Range-adaptive DBSCAN~\cite{ester1996dbscan} compensates for
inverse-square point-density decay with range. At far distances, the same
physical object subtends fewer voxels, so a fixed $\varepsilon$ would fragment
or miss clusters. The neighborhood radius adapts as:
\begin{equation}
\varepsilon(r) = \varepsilon_0 + \alpha \cdot \max(r - r_{\mathrm{ref}},\, 0),
\label{eq:adaptive_eps}
\end{equation}
where $r = \tfrac{1}{N_k}\sum_n \|\mathbf{p}_k^{(n)}\|$ is the mean scan
range, $\varepsilon_0$ is the base radius, $\alpha$ is the range-growth rate
(non-zero for adaptive configs~C and~D; zero otherwise), and $r_{\mathrm{ref}}$
is the reference distance beyond which expansion activates. Clustering assigns
labels via $\mathrm{DBSCAN}(\mathcal{P}_k^V;\, \varepsilon(r),\,
\mathrm{minPts})$, where $\mathrm{minPts}$ is the minimum point count for a
core cluster; noise-labeled points are discarded.

\subsubsection{Multi-Layer Validation}

Each candidate cluster $\mathcal{C} = \{\mathbf{p}_j\}_{j=1}^{n_c}$ passes
through three sequential validation layers.

\textit{Layer~1 -- Geometric constraints:} Point count and spatial extents
must satisfy $n_{\min} \le n_c \le n_{\max}$ and
$\max(e_x, e_y, e_z) < e_{\max}$, where $e_d$ denotes the axis-aligned
extent along axis~$d$, bounded by expected UAV dimensions.

\textit{Layer~2 -- Spatial jump filtering:} Candidates violating
$\|\mathbf{z}_k - \mathbf{z}_{k-1}\| \le \max(\tau_{\min},\, v_{\max}\Delta t)$
are rejected as physically implausible, where $\tau_{\min}$ accommodates
hovering targets and $v_{\max}$ bounds maximum velocity.

\textit{Layer~3 -- Temporal consistency:} Enabled for single-point
configurations, this layer requires at least $M$ of the last $K$ candidates
to satisfy spatial-temporal proximity:
\begin{equation}
\sum_{i=1}^{K} \mathbb{1}\bigl\{
\|\mathbf{z}_k - \mathbf{z}_{k-i}\| < d_{\mathrm{cons}}
\;\land\;
t_k - t_{k-i} < T_{\mathrm{cons}}\bigr\} \ge M.
\label{eq:local_consistency}
\end{equation}
This layer rejects sporadic noise while preserving spatially consistent sparse
returns across time.

Centroid estimation for validated clusters employs component-wise median
$\hat{\mathbf{z}}_k = [\mathrm{med}_j(p_{j,x}),\,\mathrm{med}_j(p_{j,y}),\,
\mathrm{med}_j(p_{j,z})]^\top$ when $n_c \ge 3$, providing robustness against
outlier returns from specular reflections or rotor blade hits. For sparser
clusters ($n_c < 3$), arithmetic mean is used. The resulting centroid
$\mathbf{z}_{k,j} \in \mathcal{F}_L$ is transformed to~$\mathcal{F}_G$ via
the ego-pose and passed to the tracker as a 3D position measurement.

\subsection{Multi-Object Tracking}

\subsubsection{State and Measurement Model}

Each track maintains a six-dimensional state
$\mathbf{x}_{k,i} = [\mathbf{p}_{k,i}^\top,\, \mathbf{v}_{k,i}^\top]^\top
\in \mathbb{R}^6$ in~$\mathcal{F}_G$, evolving under a constant-velocity model:
\begin{equation}
\mathbf{x}_{k|k-1} = \mathbf{F}_k\,\mathbf{x}_{k-1|k-1} + \mathbf{w}_k,
\qquad
\mathbf{F}_k =
\begin{bmatrix}
\mathbf{I}_3 & \Delta t_k\,\mathbf{I}_3 \\
\mathbf{0}   & \mathbf{I}_3
\end{bmatrix}.
\end{equation}
Process noise covariance is $\mathbf{Q}_k = \mathbf{G}_k\,q^2\mathbf{I}_3\,
\mathbf{G}_k^\top$ with gain
$\mathbf{G}_k = [\tfrac{1}{2}\Delta t_k^2\,\mathbf{I}_3;\;
\Delta t_k\,\mathbf{I}_3]^\top$. Detections provide 3D position measurements:
\begin{equation}
\mathbf{z}_{k,j} = \mathbf{H}\,\mathbf{x}_k + \mathbf{v}_k,
\qquad
\mathbf{H} = \begin{bmatrix}\mathbf{I}_3 & \mathbf{0}\end{bmatrix},
\qquad
\mathbf{v}_k \sim \mathcal{N}(\mathbf{0},\mathbf{R}).
\end{equation}

\subsubsection{IMM Filtering}

Each track employs an IMM filter~\cite{blom1988imm,li2003survey} with $M_s=3$
constant-velocity submodels at noise intensities
$q^{(m)}\!\in\!\{q_\mathrm{low}, q_\mathrm{med}, q_\mathrm{high}\}$
capturing hover, cruise, and evasive motion regimes. Model probabilities
$\mu_k^{(m)}$ evolve under Markov transition matrix $\boldsymbol{\Pi}=[\pi_{ij}]$.

\textit{Mixing:} Predicted probabilities $\mu_{k|k-1}^{(j)}\!=\!
\sum_i\pi_{ij}\mu_{k-1}^{(i)}$ yield mixing weights
$\mu_{k-1}^{(i|j)}\!=\!\pi_{ij}\mu_{k-1}^{(i)}/\mu_{k|k-1}^{(j)}$ and
mixed initial conditions per model~$j$:
\begin{equation}
\hat{\mathbf{x}}^{0(j)}_{k-1} =
\textstyle\sum_i\mu_{k-1}^{(i|j)}\hat{\mathbf{x}}^{(i)}_{k-1|k-1},
\end{equation}
\begin{equation}
\mathbf{P}^{0(j)}_{k-1} = \textstyle\sum_i\mu_{k-1}^{(i|j)}\!\left[
\mathbf{P}^{(i)}_{k-1|k-1}
+ \boldsymbol{\delta}^{(ij)}{\boldsymbol{\delta}^{(ij)}}^\top\right],
\end{equation}
where $\boldsymbol{\delta}^{(ij)}=\hat{\mathbf{x}}^{(i)}_{k-1|k-1}-
\hat{\mathbf{x}}^{0(j)}_{k-1}$.

\textit{Prediction, update, and fusion:} Each model independently predicts and
applies a standard Kalman measurement update. Model probabilities are updated
by likelihood weighting:
\begin{equation}
\mu_k^{(j)} = \frac{\Lambda_k^{(j)}\mu_{k|k-1}^{(j)}}
{\sum_l\Lambda_k^{(l)}\mu_{k|k-1}^{(l)}},
\end{equation}
where $\Lambda_k^{(j)}\!=\!\mathcal{N}(\mathbf{z}_k;\mathbf{H}
\hat{\mathbf{x}}^{(j)}_{k|k-1},\mathbf{S}_k^{(j)})$ is the innovation
likelihood. The fused state and covariance are:
\begin{equation}
\hat{\mathbf{x}}_{k|k}=\textstyle\sum_j\mu_k^{(j)}\hat{\mathbf{x}}^{(j)}_{k|k},
\quad
\mathbf{P}_{k|k}=\textstyle\sum_j\mu_k^{(j)}\!\left[\mathbf{P}^{(j)}_{k|k}
+\boldsymbol{\epsilon}^{(j)}{\boldsymbol{\epsilon}^{(j)}}^\top\right],
\end{equation}
where $\boldsymbol{\epsilon}^{(j)}=\hat{\mathbf{x}}^{(j)}_{k|k}-
\hat{\mathbf{x}}_{k|k}$. Covariance symmetrization and positive-definiteness
checks are applied after each step.

\subsubsection{Gating and Data Association}

The innovation vector and its covariance are:
\begin{equation}
\mathbf{y}_{k,j} = \mathbf{z}_{k,j} - \mathbf{H}\,\mathbf{x}_{k|k-1},
\qquad
\mathbf{S}_k = \mathbf{H}\,\mathbf{P}_{k|k-1}\,\mathbf{H}^\top + \mathbf{R}.
\end{equation}
A detection gates to a track if its squared Mahalanobis distance satisfies
$d^2(\mathbf{z}_{k,j}) = \mathbf{y}_{k,j}^\top\,\mathbf{S}_k^{-1}\,
\mathbf{y}_{k,j} \le \gamma$, where $\gamma$ is the chi-squared threshold
for three degrees of freedom. A wider gate is applied for dormant tracks to
facilitate reacquisition after prolonged detection outages. Two association
strategies are compared.

\textit{Hungarian assignment} constructs a cost matrix from gated
track-detection pairs combining Mahalanobis distance, an identity anchor
from the last confident detection, and a velocity-consistency
penalty~\cite{kuhn1955hungarian}. The optimal one-to-one assignment is
computed exactly.

\textit{Joint Probabilistic Data Association} computes marginal association
probabilities $\beta_{i,j}$ and missed-detection probabilities $\beta_{i,0}$
by enumerating all feasible joint events. The filter update uses the combined
innovation $\boldsymbol{\nu}_i = \sum_{j}\beta_{i,j}(\mathbf{z}_{k,j} -
\hat{\mathbf{z}}_{k,i})$ with inflated posterior covariance to account for
association uncertainty~\cite{bar1988tracking,fortmann1983sonar}.

\subsubsection{Track Management and Evaluation}

Tracks evolve through tentative, confirmed, and dormant states via hit/miss
counters. New tracks are initiated from high-confidence unassigned detections
subject to a minimum separation constraint to prevent duplicate initialization.
Dormant tracks are resurrected on detection of a measurement near the last
known position, enabling recovery from prolonged detection gaps. Multi-Object
Tracking Accuracy is:
\begin{equation}
\mathrm{MOTA} = 1 -
\frac{\sum_k(\mathrm{FP}_k + \mathrm{FN}_k + \mathrm{IDSW}_k)}
     {\sum_k \mathrm{GT}_k},
\end{equation}
jointly penalizing false positives, false negatives, and identity
switches~\cite{bernardin2008evaluating}.

%=============================================================================
\section{Experimental Setup}
\label{sec:setup}
%=============================================================================

The evaluation employs a two-environment protocol: real-world flights validate
detection under genuine sensor sparsity; simulation provides controlled
multi-target tracking with unambiguous identity ground truth.

\subsection{Real-World Detection Experiments}

\begin{figure}[!ht]
\centering
\includegraphics[width=0.75\columnwidth]{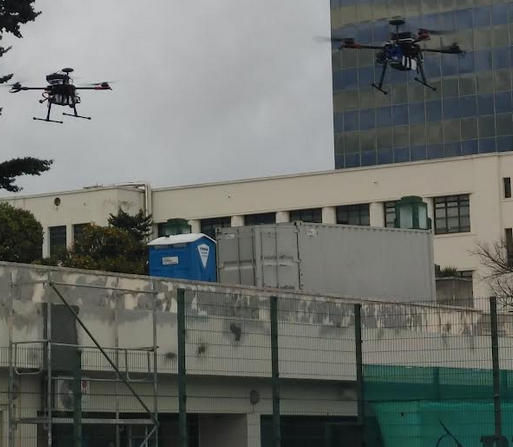}
\caption{Real-world setup: observer UAV with inverted Livox Mid-360
detecting a target quadrotor at 5--25\,m with RTK-GPS ground truth.}
\label{fig:field_test}
\end{figure}

Outdoor flight tests were conducted with two F450 quadrotors at Instituto
Superior T\'{e}cnico, Lisbon. The observer UAV carries an inverted Livox
Mid-360 solid-state LiDAR~\cite{liu2020livox} operating at 10\,Hz with
non-repetitive scanning and nominal 40\,m range. Inverted mounting biases
the field of view downward and forward for sub-observer target observation.
The target is an F450-class quadrotor (450\,mm diagonal) executing maneuvers
at 5--25\,m range. RTK-GPS provides centimeter-level relative ground truth
for both platforms (Fig.~\ref{fig:field_test}).

The dataset comprises 776 LiDAR scans over approximately 80\,s of flight;
the target was visible in 628 frames (80.9\%) under the region-of-interest
protocol. Non-repetitive scanning causes the target to intermittently exit
the instantaneous sampling region, and at ranges beyond 10\,m a 450\,mm
quadrotor typically yields only 1--2 returns per scan, making it
indistinguishable from background clutter without temporal consistency
validation.

Eight detection configurations spanning the precision-recall trade-off
are evaluated, varying $\mathrm{minPts}$, base radius $\varepsilon_0$, and
voxel size $v$ (Table~\ref{tab:det_configs}). Configurations~C and~D activate
range-adaptive growth ($\alpha>0$). All use $r_{\mathrm{ref}}=10$\,m; Layer~3
is enabled for configs~S1, S4, and~MR.

%   - Table 1: Real-world configs merged with detection results   -
\begin{table*}[!ht]
\caption{Real-World Detection Configurations and Results (628 Visible Frames,
$r_{\mathrm{ref}}=10$\,m; $^*$Layer~3 enabled).
\colorbox{rowbest}{\strut Best},
\colorbox{rowgood}{\strut viable},
\colorbox{rowbad}{\strut failure} configurations.}
\label{tab:det_configs}
\centering
\footnotesize
\setlength{\tabcolsep}{4pt}
\begin{tabular}{llcccccccccc}
\toprule
\multicolumn{5}{c}{\textbf{Configuration Parameters}} & &
\multicolumn{6}{c}{\textbf{Detection Results}} \\
\cmidrule(lr){1-5}\cmidrule(lr){7-12}
\textbf{Cfg} & \textbf{Name} & $\varepsilon_0$ (m) & \textbf{minPts} &
$v$ (m) & & \textbf{TP} & \textbf{FP} & \textbf{FN} &
\textbf{Prec} & \textbf{Rec} & \textbf{Det\%} \\
\midrule
O  & Baseline             & 0.60 & 2 & 0.05 && 46  & 1  & 582 & 0.979 & 0.073 & 7.5  \\
A  & Conservative         & 0.45 & 3 & 0.04 && 21  & 0  & 607 & 1.000 & 0.033 & 3.3  \\
\rowcolor{rowbad}
B  & Permissive           & 0.70 & 2 & 0.06 && 124 & 45 & 504 & 0.734 & 0.197 & 26.9 \\
C  & Range-Adaptive       & 0.60 & 2 & 0.04 && 45  & 1  & 583 & 0.978 & 0.072 & 7.3  \\
D  & Balanced             & 0.55 & 2 & 0.05 && 48  & 1  & 580 & 0.980 & 0.076 & 7.8  \\
\rowcolor{rowgood}
S1 & Single-Point$^*$     & 0.80 & 1 & 0.06 && 189 & 58 & 439 & 0.765 & 0.301 & \cellcolor{celltop}39.3 \\
S4 & Quad-Point$^*$       & 0.50 & 4 & 0.04 && 28  & 0  & 600 & 1.000 & 0.045 & 5.6  \\
\rowcolor{rowbest}
MR & Max Recall$^*$       & 0.80 & 2 & 0.07 && 318 & 36 & 310 &
\cellcolor{celltop}\textbf{0.891} & \cellcolor{celltop}\textbf{0.804} &
\cellcolor{celltop}\textbf{69.9} \\
\bottomrule
\end{tabular}
\vspace{1mm}
\footnotesize

\textbf{Note:} Det\% = percentage of visible frames with a detection.
Config~B: 2.30\,m RMSE (failure without geometric validation).
minPts reduction from 4 (S4) to 1 (S1) yields a sevenfold detection rate
increase (5.6\% to 39.3\%), confirming extreme per-scan sparsity.
\end{table*}

\subsection{Simulation Tracking Experiments}

\begin{figure}[!ht]
\centering
\includegraphics[width=0.89\columnwidth]{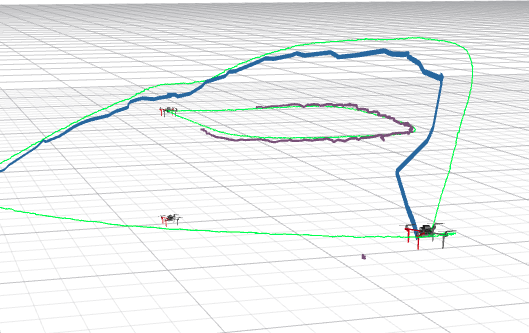}
\caption{MRS UAV System Gazebo environment for tracking evaluation with
millimeter-accurate ground truth and unambiguous identity labels.}
\label{fig:gazebo_setup}
\end{figure}

Multi-target tracking requires unambiguous identity ground truth, which GNSS
cannot provide reliably within approximately 2\,m inter-UAV separation. The
MRS UAV System~\cite{baca2021mrs} Gazebo simulator provides millimeter-accurate
pose estimates with exact identity labels and a LiDAR sensor model reproducing
the sparse-return characteristics of actual flight
(Fig.~\ref{fig:gazebo_setup}).

Four scenarios of increasing association difficulty were designed:
\begin{itemize}
  \item \textbf{Sc-I~(Occlusion, 968 frames):} Detection gaps of 3--5\,s
        test track maintenance and reacquisition.
  \item \textbf{Sc-II~(Crossings, 1708 frames):} Sustained proximity below
        3\,m with repeated trajectory crossings creates maximum association
        ambiguity.
  \item \textbf{Sc-III~(Separated, 832 frames):} Inter-target distance
        consistently above 5\,m serves as a low-ambiguity control baseline.
  \item \textbf{Sc-IV~(Moderate, 1305 frames):} Intermittent proximity at
        varying ranges represents mixed operational conditions.
\end{itemize}

Three simulation detection configurations (Table~\ref{tab:sim_configs})
combined with two association methods (Hungarian, JPDA) yield six tracker
variants per scenario (24 conditions total).

\begin{table}[!t]
\caption{Simulation Detection Configuration Parameters
(used in tracking evaluation, Sec.~\ref{sec:results}).
Voxel size is fixed by the simulation sensor model.}
\label{tab:sim_configs}
\centering
\footnotesize
\setlength{\tabcolsep}{5pt}
\begin{tabular}{llccc}
\toprule
\textbf{Cfg} & \textbf{Name} & $\varepsilon_0$ (m) & \textbf{minPts} &
\textbf{Focus} \\
\midrule
A\textsubscript{s} & Balanced  & 0.50 & 3 & Precision/recall balance \\
B\textsubscript{s} & Precision & 0.40 & 2 & Low false positives \\
C\textsubscript{s} & Recall    & 0.45 & 3 & High measurement coverage \\
\bottomrule
\end{tabular}
\end{table}

%=============================================================================
\section{Results}
\label{sec:results}
%=============================================================================

\subsection{Real-World Detection Performance}

Table~\ref{tab:det_configs} summarizes detection results across eight
configurations on 628 visible frames. A fundamental sensor constraint is
apparent throughout: the 450\,mm target produces only 1--2 returns per scan,
limiting achievable detection rate independent of parameter choice.

\subsubsection{Sparsity Quantification}

The minPts sweep directly quantifies the sparsity regime. Config.~S4
($\mathrm{minPts}=4$) achieves only 5.6\% detection rate, confirming the
target rarely provides sufficient point support for strict clustering.
Reducing to $\mathrm{minPts}=2$ (Config.~O) raises the rate to 7.5\%, while
single-point acceptance with temporal validation (Config.~S1,
$\mathrm{minPts}=1$) reaches 39.3\%, a sevenfold increase. Config.~MR
attains 69.9\% through balanced temporal validation, representing the
practical ceiling under these sensor conditions.

\subsubsection{Configuration Analysis}

\textit{Config.~MR (Max Recall)} achieves the best overall performance
(precision 0.891, recall 0.804, F1 0.845, RMSE 0.63\,m). Multi-layer
geometric and temporal validation collectively suppresses clutter while
retaining spatially consistent sparse returns, yielding a 10.2\% false
positive rate (36 of 354 detections) at 69.9\% detection coverage.

\textit{Config.~S1 (Single-Point)} demonstrates that $\mathrm{minPts}=1$
is viable under temporal consistency enforcement
(Eq.~\ref{eq:local_consistency}), achieving 0.765 precision and 0.77\,m RMSE
at 39.3\% detection rate. Sub-meter accuracy despite single-point acceptance
validates the effectiveness of the validation layers.

\textit{Config.~B (Failure Mode)} shows the consequence of parameter
permissiveness without geometric validation: an RMSE of 2.30\,m, which is
3.7 times that of Config.~MR, with 45 false positives from building edges
and pedestrian groups. This establishes that relaxing thresholds alone cannot substitute for
principled multi-layer validation.

\textit{Conservative configurations} (O, A, C, D, S4) achieve RMSE below
0.65\,m but detection rates below 8\%, suitable only for applications
tolerant of highly intermittent measurements.

\subsection{Multi-Target Tracking Performance}

Table~\ref{tab:mot_all} reports tracking results across all four scenarios,
three detection configurations, and two association methods. Results are
organized by scenario to preserve distinct ambiguity regimes; aggregate rows
summarize means across detection configurations.

\begin{table*}[!ht]
\centering
\caption{Multi-Object Tracking Results Across Scenarios, Detection
Configurations, and Association Methods.
\colorbox{rowbest}{\strut JPDA aggregate},
\colorbox{rowgood}{\strut Hung.\ aggregate},
\colorbox{rowbad}{\strut worst ID-switch cells},
\colorbox{celltop}{\strut best per-scenario MOTA}.}
\label{tab:mot_all}
\small
\setlength{\tabcolsep}{3pt}
\begin{tabular}{ll ccc ccc ccc ccc}
\toprule
& & \multicolumn{3}{c}{\textbf{Sc-I: Occlusion (968 fr.)}} &
\multicolumn{3}{c}{\textbf{Sc-II: Crossings (1708 fr.)}} &
\multicolumn{3}{c}{\textbf{Sc-III: Separated (832 fr.)}} &
\multicolumn{3}{c}{\textbf{Sc-IV: Moderate (1305 fr.)}} \\
\cmidrule(lr){3-5}\cmidrule(lr){6-8}\cmidrule(lr){9-11}\cmidrule(lr){12-14}
\textbf{Det} & \textbf{Assoc} &
\textbf{MOTA} & \textbf{RMSE} & \textbf{ID} &
\textbf{MOTA} & \textbf{RMSE} & \textbf{ID} &
\textbf{MOTA} & \textbf{RMSE} & \textbf{ID} &
\textbf{MOTA} & \textbf{RMSE} & \textbf{ID} \\
\midrule
A & Hung. & \cellcolor{celltop}0.628 & 0.624 & 4
          & \cellcolor{celltop}0.644 & 0.760 & \cellcolor{rowbad}10
          & 0.556 & 0.832 & 0
          & 0.508 & 0.500 & 1 \\
A & JPDA  & 0.615 & 0.678 & \textbf{0}
          & 0.608 & \textbf{0.716} & 5
          & 0.556 & 0.842 & \textbf{0}
          & \cellcolor{celltop}0.545 & 0.579 & \textbf{0} \\
\midrule
B & Hung. & 0.626 & 0.712 & \cellcolor{rowbad}7
          & 0.603 & 0.793 & \cellcolor{rowbad}12
          & 0.491 & 0.585 & 0
          & 0.506 & 0.511 & 1 \\
B & JPDA  & 0.602 & 0.711 & \textbf{0}
          & 0.601 & 0.741 & 7
          & 0.482 & 0.583 & \textbf{0}
          & 0.529 & 0.487 & \textbf{0} \\
\midrule
C & Hung. & 0.626 & 0.657 & \cellcolor{rowbad}5
          & 0.640 & 0.778 & \cellcolor{rowbad}12
          & 0.536 & 0.838 & 0
          & 0.507 & 0.520 & 1 \\
C & JPDA  & 0.612 & 0.716 & \textbf{0}
          & 0.606 & 0.740 & 7
          & \cellcolor{celltop}0.544 & 0.841 & \textbf{0}
          & 0.544 & 0.600 & \textbf{0} \\
\midrule
\multicolumn{14}{c}{\textit{Aggregate means across detection configurations}} \\
\midrule
\rowcolor{rowgood}
-- & Hung. & 0.627 & 0.664 & \cellcolor{rowbad}5.3
           & 0.629 & 0.777 & \cellcolor{rowbad}11.3
           & 0.528 & 0.752 & 0
           & 0.507 & 0.510 & 1.0 \\
\rowcolor{rowbest}
-- & JPDA  & 0.610 & 0.702 & \textbf{0}
           & 0.605 & \textbf{0.732} & \textbf{6.3}
           & 0.527 & 0.755 & \textbf{0}
           & \textbf{0.539} & 0.555 & \textbf{0} \\
\bottomrule
\end{tabular}
\vspace{1mm}
\footnotesize

\textbf{Note:} ID = identity switches; RMSE in meters.
JPDA reduces switches by 64\% (mean 4.4 to 1.6 across Sc-I, Sc-II, Sc-IV)
at a MOTA cost of 0.003.
\end{table*}

\subsubsection{Per-Scenario Analysis}

\textit{Sc-I (Occlusion).}
Prolonged detection gaps of 3--5\,s expose differences in track reacquisition.
Hungarian achieves the highest per-scenario MOTA of 0.628
(Config.~A\textsubscript{s}) but accumulates 4--7 identity switches across
detection configurations. JPDA eliminates switches entirely under
Configs.~A\textsubscript{s} and C\textsubscript{s}, demonstrating robustness
to sustained detection outages through probabilistic track maintenance.

\textit{Sc-II (Crossings).}
Sustained proximity below 3\,m with repeated crossings produces maximum
association ambiguity. Hungarian achieves 0.644 MOTA
(Config.~A\textsubscript{s}) but incurs 10--12 identity switches across
all configurations. JPDA reduces switches to 5--7 and yields lower
localization error (0.716\,m versus 0.760\,m under Config.~A\textsubscript{s}),
consistent with soft-assignment updating mitigating hard-commitment errors
during close-range crossing events.

\textit{Sc-III (Separated).}
With inter-target separation consistently above 5\,m, both methods yield
zero identity switches across all configurations. This confirms that
performance differences in Sc-I and Sc-II are attributable to association
ambiguity rather than filter instability or motion model mismatch.

\begin{figure}[!ht]
\centering
\includegraphics[width=0.79\columnwidth]{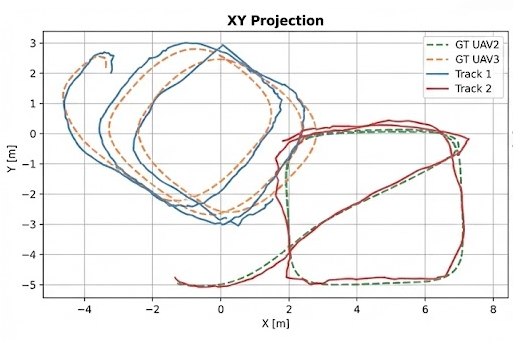}
\caption{XY trajectory projection for Sc-IV (Moderate). Ground-truth paths
(dashed) and JPDA track estimates (solid) are overlaid for both UAVs.
JPDA successfully maintains correct track identities throughout intermittent
proximity events, demonstrating its advantage over Hungarian assignment in
mixed-ambiguity operational conditions.}
\label{fig:xy_s4}
\end{figure}

\textit{Sc-IV (Moderate).}
Even intermittent proximity benefits probabilistic association. JPDA achieves
higher MOTA (0.545 versus 0.508 under Config.~A\textsubscript{s}) with zero
identity switches, indicating that soft assignment is advantageous beyond
exclusively sustained-crossing conditions. Fig.~\ref{fig:xy_s4} illustrates
this scenario: JPDA maintains correct track identities throughout the
intermittent proximity events, whereas Hungarian assignment incurs one switch
under the same conditions.

\subsubsection{Detection Configuration Impact on Tracking}

Config.~A\textsubscript{s} (Balanced) yields the highest mean MOTA for
Hungarian (0.584 across all scenarios), as balanced precision-recall provides
consistent measurement input. Config.~C\textsubscript{s} (Recall-biased)
benefits JPDA most, enabling 0.570 mean MOTA with zero switches in Sc-I and
Sc-IV; increased measurement availability amplifies JPDA's probabilistic
advantage. Config.~B\textsubscript{s} (Precision-biased) produces fewest
identity switches for both methods but lower mean MOTA (0.551 Hungarian,
0.554 JPDA), as missed detections during critical periods reduce tracking
continuity. This detection-association coupling is a key design consideration:
recall-biased configurations enhance JPDA through availability, while
precision-biased configurations reduce ambiguity at the cost of coverage.

\subsubsection{Aggregated Performance}

Aggregating across Sc-I, Sc-II, and Sc-IV (Sc-III excluded as both methods
yield zero switches) over all detection configurations, JPDA reduces identity
switches by 64\% (mean 4.4 to 1.6) at a MOTA cost of 0.003 (0.573 to 0.570).
Localization error increases marginally by 1.5\% (0.676\,m to 0.686\,m).
These results quantify the accuracy-identity trade-off: deterministic
assignment favors geometric fidelity through hard measurement commitment,
while probabilistic association prioritizes identity preservation under
ambiguous interactions at negligible accuracy cost.

%=============================================================================
\section{Discussion}
\label{sec:discussion}
%=============================================================================

\subsection{Detection Under Extreme Sparsity}

The 69.9\% detection ceiling despite 80.9\% target visibility reflects the
intrinsic sensor constraint: at 10--25\,m with rosette scanning, a 450\,mm
quadrotor yields only 1--2 returns per scan, well below the 20+ points
assumed by automotive 3D detectors at
70\,m~\cite{qi2018frustum,lang2019pointpillars}. The sevenfold rate increase
from Config.~S4 to Config.~S1 confirms that minPts is the dominant operational
parameter and that most scans provide insufficient point support.

Config.~MR shows that temporal multi-layer validation enables reliable
detection despite per-scan insufficiency; Config.~B's 2.30\,m RMSE confirms
that permissiveness without geometric validation degrades localization
regardless of recall level. Measurement quality must therefore be prioritized
over raw detection rate at the detector-tracker interface.
This contrasts with~\cite{khosravi2026lightweight3dlidarbaseduav}, which
addresses LiDAR dropout via adaptive EKF with motion-model probability
switching, without inertial measurements. The two approaches are
complementary, targeting different pipeline stages.

\subsection{Data Association Performance}

JPDA's 64\% identity-switch reduction at 0.003 MOTA cost confirms that
probabilistic association improves identity preservation under close-proximity
ambiguity at negligible accuracy cost. Gains are scenario-dependent:
both methods perform equivalently under separation (Sc-III), while JPDA
advantages emerge clearly during crossings (Sc-II) and intermittent proximity
(Sc-IV). Hungarian achieves marginally higher instantaneous MOTA through
aggressive geometric fitting, but hard commitment during ambiguous interactions
accumulates identity errors costly for cooperative localization and collision
avoidance. JPDA's modest RMSE increase (0.676\,m to 0.686\,m) is a favorable
exchange when long-term track continuity is required.

\subsection{System Implications and Limitations}

Config.~MR or Config.~S1 is recommended for state estimation; Hungarian is
appropriate for well-separated targets; JPDA is preferred whenever identity
continuity matters during close maneuvers. Temporal consistency validation
is essential; threshold tuning alone is insufficient, as Config.~B confirms.

Limitations include single-target real-world validation; reliable multi-UAV
ground truth at close range requires UWB-based
localization~\cite{mueller2013quadcopter} or motion capture. JPDA scales
combinatorially and requires JIPDA~\cite{musicki2004integrated} for swarms.
IMM parameters require recalibration for different platform dynamics.

%=============================================================================
\section{Conclusion}
\label{sec:conclusion}
%=============================================================================

Solid-state LiDAR imposes a hard sparsity floor in airborne UAV perception:
at 10--25\,m, a 450\,mm quadrotor yields only 1--2 returns per scan due to
non-repetitive rosette scanning, irrespective of parameter choice. This work
showed that reliable operation within this regime requires co-design of
detector and tracker rather than threshold tuning alone.

On detection, three-stage temporal consistency validation achieves 0.891
precision and 0.804 recall with 0.63\,m RMSE, approaching the physical
69.9\% detection ceiling. Permissive clustering without geometric validation
degrades localization to 2.30\,m RMSE, confirming that measurement quality
dominates over raw detection rate at the detector-tracker interface.

On tracking, JPDA reduces identity switches by 64\% over Hungarian at
0.003 MOTA cost. Gains concentrate in crossing and intermittent-proximity
scenarios and are amplified by recall-biased detection, establishing a
detector-tracker coupling: high-recall detection feeds more measurements
to JPDA, directly amplifying its probabilistic advantage.

The two-environment protocol, combining real-world RTK-GPS detection
validation with simulation-based identity-labeled tracking, provides a sound
methodology for systems where GNSS ground truth is insufficient at close
inter-UAV separations. Future work will extend to per-layer ablation studies, real-world
multi-UAV tracking with UWB-based relative localization, and swarm-scale
JIPDA approximations.

\section*{Acknowledgment}
The authors acknowledge the financial support provided by the Aero.Next Project under Grant Nos.\ C645727867 and 00000066.

\balance
\bibliographystyle{IEEEtran}
\bibliography{ref}

@inproceedings{ritchie2015micro,
  author    = {Ritchie, M. and Fioranelli, F. and Griffiths, H. and Torvik, B.},
  title     = {Micro-drone {RCS} analysis},
  booktitle = {Proc. {IEEE} Radar Conf. (RadarConf)},
  year      = {2015},
  pages     = {452--456},
  doi       = {10.1109/RadarConf.2015.7411926}
}

@inproceedings{mueller2013quadcopter,
  author    = {Mueller, M. W. and Hamer, M. and {D'Andrea}, R.},
  title     = {Fusing ultra-wideband range measurements with accelerometers and rate gyroscopes for quadrocopter state estimation},
  booktitle = {Proc. {IEEE} Int. Conf. Robot. Autom. (ICRA)},
  year      = {2015},
  pages     = {1730--1736},
  doi       = {10.1109/ICRA.2015.7139421}
}

@inproceedings{nguyen2018drone,
  author    = {Nguyen, P. and Ravindranatha, M. and Nguyen, A. and Han, R. and Vu, T.},
  title     = {Investigating cost-effective {RF}-based detection of drones},
  booktitle = {Proc. {ACM} DroNet Workshop},
  year      = {2016},
  pages     = {17--22},
  doi       = {10.1145/3001867.3001868}
}

@article{ezuma2019detection,
  author  = {Ezuma, M. and Erden, F. and Anjinappa, C. K. and Ozdemir, O. and Guvenc, I.},
  title   = {Detection and classification of {UAV}s using {RF} fingerprints in the presence of {Wi-Fi} and {Bluetooth} interference},
  journal = {{IEEE} Open Journal of the Communications Society},
  volume  = {1},
  pages   = {60--76},
  year    = {2020}
}

@article{basiri2016sound,
  author  = {Basiri, M. and Schill, F. and Lima, P. and Floreano, D.},
  title   = {On-board relative bearing estimation for teams of drones using sound},
  journal = {{IEEE} Robotics and Automation Letters},
  volume  = {1},
  number  = {2},
  pages   = {820--827},
  month   = jul,
  year    = {2016}
}

@article{rozantsev2017detecting,
  author  = {Rozantsev, A. and Lepetit, V. and Fua, P.},
  title   = {Detecting flying objects using a single moving camera},
  journal = {{IEEE} Trans. Pattern Anal. Mach. Intell.},
  volume  = {39},
  number  = {5},
  pages   = {879--892},
  year    = {2017}
}

@inproceedings{bjorklund2018mutual,
  author    = {Bj{\"o}rklund, S. and {\"O}hman, A.},
  title     = {Target detection and classification of small drones by deep learning on radar micro-{Doppler}},
  booktitle = {Proc. European Radar Conf. (EuRAD)},
  year      = {2018},
  pages     = {182--185}
}

@inproceedings{hammer2018lidar,
  author    = {Hammer, M. and Hebel, M. and Laurenzis, M. and Arens, M.},
  title     = {{LiDAR}-based detection and tracking of small {UAV}s},
  booktitle = {Proc. SPIE},
  volume    = {10799},
  year      = {2018},
  note      = {Paper 107990S}
}

@inproceedings{bewley2016simple,
  author    = {Bewley, A. and Ge, Z. and Ott, L. and Ramos, F. and Upcroft, B.},
  title     = {Simple online and realtime tracking},
  booktitle = {Proc. {IEEE} Int. Conf. Image Process. (ICIP)},
  year      = {2016},
  pages     = {3464--3468}
}

@inproceedings{wojke2017simple,
  author    = {Wojke, N. and Bewley, A. and Paulus, D.},
  title     = {Simple online and realtime tracking with a deep association metric},
  booktitle = {Proc. {IEEE} Int. Conf. Image Process. (ICIP)},
  year      = {2017},
  pages     = {3645--3649}
}

@book{bar1988tracking,
  author    = {Bar-Shalom, Y. and Fortmann, T. E.},
  title     = {Tracking and Data Association},
  publisher = {Academic Press},
  address   = {San Diego, CA, USA},
  year      = {1988}
}

@inproceedings{ester1996dbscan,
  author    = {Ester, M. and Kriegel, H.-P. and Sander, J. and Xu, X.},
  title     = {A density-based algorithm for discovering clusters in large spatial databases with noise},
  booktitle = {Proc. Int. Conf. Knowledge Discovery and Data Mining (KDD)},
  year      = {1996},
  pages     = {226--231}
}

@inproceedings{qi2017pointnet,
  author    = {Qi, C. R. and Su, H. and Mo, K. and Guibas, L. J.},
  title     = {{PointNet}: Deep learning on point sets for {3D} classification and segmentation},
  booktitle = {Proc. {IEEE} Conf. Comput. Vis. Pattern Recognit. (CVPR)},
  year      = {2017},
  pages     = {652--660}
}

@inproceedings{zhou2018voxelnet,
  author    = {Zhou, Y. and Tuzel, O.},
  title     = {{VoxelNet}: End-to-end learning for point cloud based {3D} object detection},
  booktitle = {Proc. {IEEE} Conf. Comput. Vis. Pattern Recognit. (CVPR)},
  year      = {2018},
  pages     = {4490--4499}
}

@article{dogru2022drone,
  author  = {Dogru, S. and Marques, L.},
  title   = {Drone detection using sparse {LiDAR} measurements},
  journal = {{IEEE} Robotics and Automation Letters},
  volume  = {7},
  number  = {2},
  pages   = {3062--3069},
  month   = apr,
  year    = {2022},
  doi     = {10.1109/LRA.2022.3145498}
}

@inproceedings{razlaw2019detection,
  author    = {Razlaw, J. and Quiskamp, D. and Burschka, D. and Lauer, M.},
  title     = {Detection and tracking of small {UAV}s in dense urban environments},
  booktitle = {Proc. {IEEE} Int. Conf. Robot. Autom. (ICRA)},
  year      = {2019},
  pages     = {1266--1272}
}

@misc{liang2024unsupervised,
  author       = {Liang, H. and Cao, S. and Wang, Z. and Liu, J.},
  title        = {Unsupervised {UAV} {3D} trajectories estimation with sparse point clouds},
  howpublished = {arXiv preprint},
  year         = {2024},
  eprint       = {2412.12716},
  archivePrefix= {arXiv}
}

@article{liu2020livox,
  author  = {Liu, Z. and Zhang, F. and Hong, X.},
  title   = {Low-cost retina-like robotic {LiDAR} for {3D} dense depth sensing},
  journal = {Sensors},
  volume  = {20},
  number  = {9},
  pages   = {2582},
  year    = {2020},
  doi     = {10.3390/s20092582}
}

@article{kuhn1955hungarian,
  author  = {Kuhn, H. W.},
  title   = {The {Hungarian} method for the assignment problem},
  journal = {Naval Research Logistics Quarterly},
  volume  = {2},
  number  = {1-2},
  pages   = {83--97},
  year    = {1955}
}

@article{fortmann1983sonar,
  author  = {Fortmann, T. and Bar-Shalom, Y. and Scheffe, M.},
  title   = {Sonar tracking of multiple targets using joint probabilistic data association},
  journal = {{IEEE} Journal of Oceanic Engineering},
  volume  = {8},
  number  = {3},
  pages   = {173--184},
  month   = jul,
  year    = {1983}
}

@inproceedings{rezatofighi2015joint,
  author    = {Rezatofighi, S. H. and Milan, A. and Zhang, Z. and Shi, Q. and Dick, A. and Reid, I.},
  title     = {Joint probabilistic data association revisited},
  booktitle = {Proc. {IEEE} Int. Conf. Comput. Vis. (ICCV)},
  year      = {2015},
  pages     = {3047--3055}
}

@article{musicki2004integrated,
  author  = {Musicki, D. and Evans, R. and Stankovi{\'c}, S.},
  title   = {Integrated probabilistic data association},
  journal = {{IEEE} Trans. Autom. Control},
  volume  = {39},
  number  = {6},
  pages   = {1237--1241},
  month   = jun,
  year    = {1994}
}

@article{blom1988imm,
  author  = {Blom, H. A. P. and Bar-Shalom, Y.},
  title   = {The interacting multiple model algorithm for systems with {Markovian} switching coefficients},
  journal = {{IEEE} Trans. Autom. Control},
  volume  = {33},
  number  = {8},
  pages   = {780--783},
  month   = aug,
  year    = {1988}
}

@article{li2003survey,
  author  = {Li, X. R. and Jilkov, V. P.},
  title   = {Survey of maneuvering target tracking. {Part I}: Dynamic models},
  journal = {{IEEE} Trans. Aerosp. Electron. Syst.},
  volume  = {39},
  number  = {4},
  pages   = {1333--1364},
  month   = oct,
  year    = {2003}
}

@inproceedings{weng2020ab3dmot,
  author    = {Weng, X. and Wang, J. and Held, D. and Kitani, K.},
  title     = {{AB3DMOT}: A baseline for {3D} multi-object tracking and new evaluation metrics},
  booktitle = {Proc. {IEEE/RSJ} Int. Conf. Intelligent Robots and Systems (IROS)},
  year      = {2020},
  pages     = {10359--10366},
  doi       = {10.1109/IROS45743.2020.9341164}
}

@inproceedings{chiu2021probabilistic,
  author    = {Chiu, H.-K. and Prioletti, A. and Li, J. and Bohg, J.},
  title     = {Probabilistic {3D} multi-modal, multi-object tracking for autonomous driving},
  booktitle = {Proc. {IEEE} Int. Conf. Robot. Autom. (ICRA)},
  year      = {2021},
  pages     = {14227--14233}
}

@article{baca2021mrs,
  author  = {Baca, T. and Petrlik, M. and Vrba, M. and Spurny, V. and Penicka, R. and Hert, D. and Saska, M.},
  title   = {The {MRS} {UAV} system: Pushing the frontiers of reproducible research, real-world deployment, and education with autonomous unmanned aerial vehicles},
  journal = {Journal of Intelligent \& Robotic Systems},
  volume  = {102},
  number  = {26},
  pages   = {1--28},
  year    = {2021},
  doi     = {10.1007/s10846-021-01383-5}
}

@article{bernardin2008evaluating,
  author  = {Bernardin, K. and Stiefelhagen, R.},
  title   = {Evaluating multiple object tracking performance: The {CLEAR MOT} metrics},
  journal = {EURASIP Journal on Image and Video Processing},
  volume  = {2008},
  pages   = {1--10},
  year    = {2008}
}

@misc{khosravi2026lightweight3dlidarbaseduav,
  title   = {Lightweight {3D} {LiDAR}-Based {UAV} Tracking: An Adaptive
             Extended {Kalman} Filtering Approach},
  author  = {Khosravi, Nivand and Basiri, Meysam and Ventura, Rodrigo},
  year    = {2026},
  eprint  = {2603.09783},
  archivePrefix = {arXiv},
  primaryClass  = {cs.RO},
  url     = {https://arxiv.org/abs/2603.09783}
}

@article{holz2015registration,
  author  = {Holz, Dirk and Ichim, Alexandru E. and Tombari, Federico and Rusu, Radu B. and Behnke, Sven},
  title   = {Registration with the Point Cloud Library: A Modular Framework for Aligning in 3-D},
  journal = {IEEE Robotics \& Automation Magazine},
  volume  = {22},
  number  = {4},
  pages   = {110--124},
  year    = {2015}
}

@inproceedings{qi2017pointnetpp,
  author    = {Qi, Charles R. and Yi, Li and Su, Hao and Guibas, Leonidas J.},
  title     = {{PointNet++}: Deep Hierarchical Feature Learning on Point
               Sets in a Metric Space},
  booktitle = {Advances in Neural Information Processing Systems (NeurIPS)},
  volume    = {30},
  year      = {2017}
}

@inproceedings{lang2019pointpillars,
  author    = {Lang, Alex H. and Vora, Sourabh and Caesar, Holger
               and Zhou, Lubing and Yang, Jiong and Beijbom, Oscar},
  title     = {{PointPillars}: Fast Encoders for Object Detection from
               Point Clouds},
  booktitle = {Proc.\ IEEE/CVF Conf.\ Comput.\ Vis.\ Pattern Recognit.\ (CVPR)},
  pages     = {12697--12705},
  year      = {2019},
  publisher = {IEEE}
}

@inproceedings{shi2019pointrcnn,
  author    = {Shi, Shaoshuai and Wang, Xiaogang and Li, Hongsheng},
  title     = {{PointRCNN}: {3D} Object Proposal Generation and Detection
               from Point Cloud},
  booktitle = {Proc.\ IEEE/CVF Conf.\ Comput.\ Vis.\ Pattern Recognit.\ (CVPR)},
  pages     = {770--779},
  year      = {2019},
  publisher = {IEEE}
}

@inproceedings{qi2018frustum,
  author    = {Qi, Charles R. and Liu, Wei and Wu, Chenxia
               and Su, Hao and Guibas, Leonidas J.},
  title     = {Frustum {PointNets} for {3D} Object Detection from
               {RGB-D} Data},
  booktitle = {Proc.\ IEEE Conf.\ Comput.\ Vis.\ Pattern Recognit.\ (CVPR)},
  pages     = {918--927},
  year      = {2018},
  publisher = {IEEE}
}

@inproceedings{rusu20113d,
  author    = {Rusu, Radu Bogdan and Cousins, Steve},
  title     = {{3D} is Here: {Point Cloud Library} ({PCL})},
  booktitle = {Proc.\ IEEE Int.\ Conf.\ Robot.\ Autom.\ (ICRA)},
  pages     = {1--4},
  year      = {2011},
  publisher = {IEEE}
}
\end{document}